\documentclass[10pt,twocolumn,letterpaper]{article}

\usepackage{textcomp}
\usepackage{wacv}
\usepackage{times}
\usepackage{epsfig}
\usepackage{graphicx}
\usepackage{amsmath}
\usepackage{amssymb}
\usepackage{url}
\usepackage{verbatim}
\usepackage{adjustbox}
\usepackage{changepage}
\usepackage{multirow}
\usepackage{xcolor}
\usepackage{caption}
\usepackage{amsmath,array,graphicx}
\usepackage[font=small,labelfont=bf]{caption}

\wacvfinalcopy 

\def\wacvPaperID{1303} 

\ifwacvfinal\fi
\begin{document}

\title{DeFraudNet:End2End Fingerprint Spoof Detection using Patch Level Attention }


\author{B.V.S Anusha \\
IIT Bombay \\
{\tt\small bvs.anusha26@gmail.com}
\and
Sayan Banerjee \\
IIT Bombay\\
{\tt\small sayan91.ban@gmail.com}
\and
Subhasis Chaudhuri \\
IIT Bombay\\
{\tt\small sc@ee.iitb.ac.in}
}

\maketitle
\ifwacvfinal\thispagestyle{empty}\fi

\begin{abstract}
In recent years, fingerprint recognition systems have made remarkable advancements in the field of biometric security as it plays an important role in personal, national and global security. In spite of all these notable advancements, the fingerprint recognition technology is still susceptible to spoof attacks which can significantly jeopardize the user security. The cross sensor and cross material spoof detection still pose a challenge with a myriad of spoof materials emerging every day, compromising sensor interoperability and robustness. This paper proposes a novel method for fingerprint spoof detection using both global and local fingerprint feature descriptors. These descriptors are extracted using DenseNet which significantly improves cross-sensor, cross-material and cross-dataset performance. A novel patch attention network is used for finding the most discriminative patches and also for network fusion. We evaluate our method on four publicly available datasets: LivDet 2011, 2013, 2015 and 2017. A set of comprehensive experiments are carried out to evaluate cross-sensor, cross-material and cross-dataset performance over these datasets. The proposed approach achieves an average accuracy of \textbf{99.52\%, 99.16\%} and \textbf{99.72\%} on LivDet 2017, 2015 and 2011 respectively outperforming the current state-of-the-art results by \textbf{3\%} and \textbf{4\%} for LivDet 2015 and 2011 respectively.   
\end{abstract}
\vspace{-0.6cm}
\section{Introduction}
Over the past few years, with an exponential increase in the usage of IoT devices, biometrics have played a key role in maintaining user confidentiality for various operations. According to a report \cite{bayometric1,bayometric2}, the fingerprint is the most widely used biometric identity over any other existing biometrics. It's extensive global usage makes it vulnerable to several security threats like, identity theft, account hacking, unauthorized access and many more. One of the many threats that can severely compromise fingerprint security is fingerprint spoofing. It is a technique in which fake fingerprint impressions are created to fool the fingerprint sensor to make unauthorized access into the fingerprint system.

\begin{figure}[t]   
\centering
  \includegraphics[scale = 0.32]{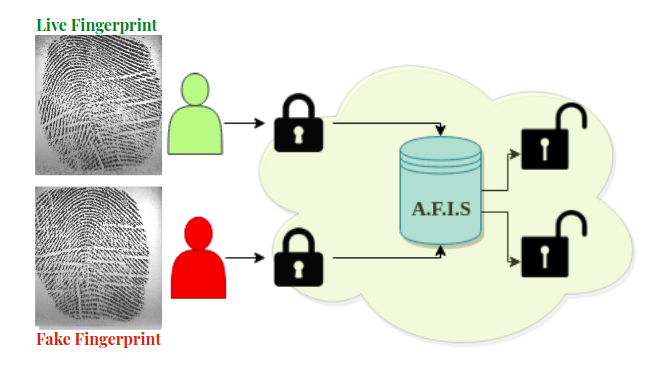}
  
  \begin{center}
    \caption{(a)}
  \end{center}
  \includegraphics[scale = 0.35]{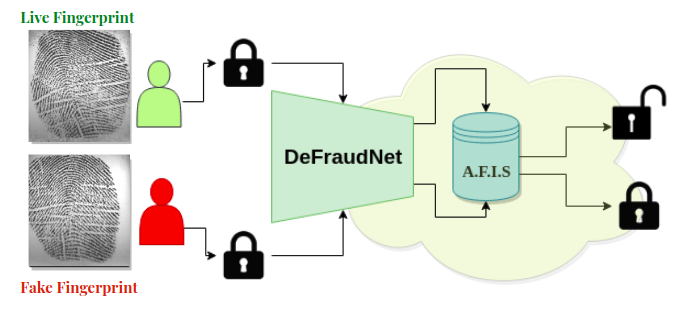}
   \begin{center}
    \caption{(b)}
  \end{center}
  
  \caption{(a) Any Automatic Fingerprint Identification System (A.F.I.S) without spoof detection system is susceptible to identity theft as the Fake and Live fingerprint have similar properties. (b) The presence of Spoof Detection Systems like DeFraudNet, prevents such threats by filtering out the Fake fingerprints and allow only the Live fingerprints into the A.F.I.S.}
  \vspace{-0.5cm}
\end{figure}

To overcome this problem, over the years various spoof detection methods have been developed. But as the fingerprint sensing technology advances, so does the spoofing technology, which increases the degree of difficulty for organizations to protect their biometric systems from being compromised. The spoof fingerprints can be fabricated using various materials like latex, ecoflex, clay, wood glue, gum etc. Visually, no clear distinction can be observed between the spoofed and live fingerprint on the sensor imagery as shown in Figure 1. Therefore, it is necessary to extract textural, anatomical or physiological features for spoof detection. 

Traditional fingerprint spoof detection methods \cite{abhyankar2006fingerprint,nikam2008fingerprint,akhtar2015correlation,gottschlich2014fingerprint} extract handcrafted texture features and use them to classify fingerprints into live/spoof classes. These methods require high resolution images and exhaustive feature tuning. As a result, they become sensitive to the computed features and input noise. To address this challenge, Menotti \textit{et al.} developed a CNN based network called '\textit{spoofnet}' \cite{menotti2015deep} and trained the network on LivDet 2013 dataset. This supervised network learns robust high level features which improves the performance by a significant margin. Following the same path, Nouguira \textit{et al.} \cite{article} and Pala \textit{et al.} \cite{Pala2017} applied standard CNN networks (VGGNet, AlexNet) pre-trained on large ImageNet dataset and fine tuned on LiveDet dataset. Use of a pre-trained network with transfer learning further enhances the performance. Inspired by these works, a lot of CNN based fingerprint spoof detection methods have been proposed \cite{article,He2015DeepRL,Woo_2018_ECCV,Simonyan14verydeep,chollet2016xception}. However, even though these methods provide affirmative results on fingerprint spoof detection on a single sensor data, their cross material or cross sensor performance is very poor. 

To improve the performance of spoof detection across cross-material and cross-sensor, we propose a novel CNN based end-to-end model which takes global images and the corresponding overlapping local patches to classify the fingerprints as live or fake. It should be noted that T. Chugh \textit{et al.} in their paper \cite{Chugh2017FingerprintSB} also used fingerprint patches but they are computed using significant minutiae points. This method obtains significantly improved results but has a few drawbacks. Firstly, minutiae point extraction requires high resolution input fingerprint images ($>$ 500dpi) and secondly, due to this preprocessing step, the model does not operate in end-to-end fashion. Taking these problems into consideration, the proposed method extracts suitable patches implicitly using a novel attention mechanism, referred as patch attention, along with channel attention and spatial attention modules. The role of the channel and spatial attention modules are to highlight useful information from each patch for live/spoof detection which together helps the patch attention module to identify appropriate patches from a pool of patches. The salient part is that, all of these attentions are learned using different neural networks with back propagation of the gradient of the main loss function. Hence, the proposed patch extraction method can be integrated with any neural network based system. DenseNet has been used as the base network for feature computation of the input image and patches. It is lighter in terms of memory usage than the existing state-of-the-art networks even with the presence of attention modules. One of the most challenging part for any patch based method is to develop an efficient fusion strategy to integrate information obtained across individual patches. Instead of using the standard approaches which predominantly use strategies like majority voting the proposed method uses patch attention network. It first determines importance of each patch to the final objective of live/spoof classification and then aggregates this information based upon their individual importance scores. By this process, the network itself learns to perform better decision fusion which also indirectly helps to learn better features for identifying spoof fingerprints. 

Extensive experiments and ablation studies have been performed on various challenging fingerprint spoof detection datasets: LivDet 2011,2013,2015,2017, which demonstrate significant improvement over state-of-the-art methods.
The main contributions of this paper are:
\begin{itemize}
    \item With the use of local patch features and global contextual image features, the proposed method obtains significantly better performance than the existing state-of-the-art fingerprint spoof detection methods across cross material and cross datasets.
    \item The proposed method also exploits handcrafted features (i.e. LBP and Gabor filters) which are integrated along with the input image. The combination of handcrafted features and deep high level semantic features show a significant improvement over cross material and cross datasets.  
    \item The proposed novel patch attention network learns highly discriminative patches with additional channel and spatial attention modules using gradients of the live/spoof classification error. Therefore unlike the existing methods, the proposed method does not require any separate intermediate step for patch discrimination. It learns to identify useful patches by itself and the complete network can be operated in an end-to-end manner.
    \item The proposed model employs a novel feature fusion strategy using patch attention. It learns to aggregate information across patches which in turn makes it less susceptible to input noise and error in initial patch computation.  
    \item The proposed network reduces the memory usage by at least fifty percentage as compared to the state-of-the-art networks. Therefore, it can be easily embedded into tiny low powered, low memory IoT devices (for example: mobile phones). 
    
\end{itemize}

\section{Related Work}
In this section, we briefly review the existing work on fingerprint spoof detection, image classification using DenseNet and use of attention network for binary classification.
\\\textbf{Fingerprint spoof detection} has been extensively studied and experimented over the centuries as it poses a huge threat to security. The spoof detection methods can be categorised into hardware based methods and software based methods. Hardware based methods involve external fingerprint sensing devices by adding sensors to detect living traits like blood pressure, blood sugar, skin distortion or odor \cite{tan2008new,choi2007aliveness}. The software based methods involve extracting various handcrafted features from the sensor image of fingerprint and then classify them as live/fake. These handcrafted features can be broadly classified as outer anatomical features like ridge strength, pore location and their distribution, continuity and clarity \cite{Pores} etc. or physiological features like perspiration patterns \cite{Percipiration} and texture based features or statistical features like Weber Local Descriptors (WLD) \cite{WLD}, rotation invariant Local Phase Quantization (LPQ) \cite{LPQ} features or Binarized Statistical Image Features (BSIF) \cite{BSIF} or Local Binary Patterns \cite{Ojala1994PerformanceEO} etc. or a combination of these features. We use textural features (i.e LBP and Gabor filters) in our approach.
LBP \cite{Ojala1994PerformanceEO} provides good textural variation for liveliness detection.
In 2015 LiveDet competition, Noguira \textit{et al.} \cite{article} obtained a state-of-the-art accuracies by using LBP and transfer learning for binary classification. In this method, the fingerprint data was preprocessed by extracting LBP features and then classified using pre-trained standard networks VGG, AlexNet. Pala \textit{et al.} \cite{Pala2017} and Menotti \textit{et al.} \cite{menotti2015deep} also use similar methods for spoof results and obtain significantly better results than spoof detection using only handcrafted features. However, all of these methods perform poorly on cross-sensor and cross material tests. To overcome this, T.Chugh \textit{et al.} \cite{Chugh2017FingerprintSB} developed a robust spoof detection method using local minutiae based patches and trained them using MobileNet-v1 \cite{Howard2017MobileNetsEC}. They obtained state-of-the-art results for intra-sensor, inter-sensor, cross-material and cross sensor over three datasets (LivDet 2011,2013,2015). But their method was not end to end as it involved two stage training process.
\\\textbf{DenseNet} \cite{Huang2016Densely} is being extensively used for various applications like image classification \cite{zhang2019channel,zhang2019multiple}, segmentation \cite{yuan2019prostate}, image super-resolution \cite{wang2018reconstructed} etc. This can be attributed to its memory efficiency, computational efficiency and feature re-usability properties. This network alleviates vanishing gradient problem and strengthens feature propagation. It's feature reusability ensures memory and computational efficiency. George \textit{et al.} in their paper \cite{george2019deep} used DenseNet to counterfeit presentation attack on human faces. Huang \textit{et al.} \cite{huang2019audio} employed DenseNet along with LSTM for audio spoof attack detection. But with the best of our knowledge, DenseNet has not been used for fingerprint spoof detection.
\\\textbf{Channel and Spatial Attention networks} highlights the salient features from visual data \cite{wang2017residual,chen2017sca,wang2019salient}. Attention has been used for face anti-spoofing \cite{chen2019attention} but as far as we know, attention has never been exploited for fingerprint spoof detection. In this paper we use attention in a similar way as mentioned in \cite{Woo_2018_ECCV}. 
\\\textbf{Patch Attention Network} is a novel attention module proposed in this paper, which learns the most discriminative patches from a set of input patches. The patch-based training always poses a problem of fusing the patch-level predictions considering the fact that not all patches are equally informative. Le Hou \textit{et al.} \cite{7780635} overcame this challenge by using a novel Expectation-Maximization (EM) based method that automatically locates discriminative patches robustly by utilizing the inter-spatial relationship of patches. 
\begin{figure*}[htb]
\begin{center}
\fbox{ \includegraphics[width = 0.8\linewidth]{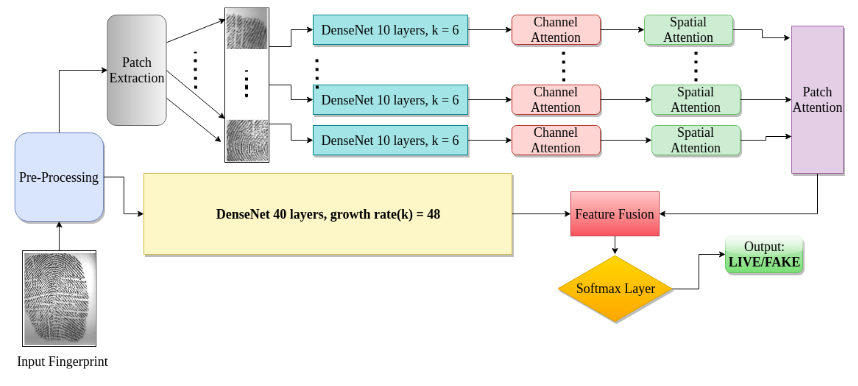}}
\end{center}
   \caption{Complete overview of the proposed model. It should be noted that in this figure the dotted lines signify continuation of the blocks.}
\label{fig:short}
\end{figure*}
\section{Proposed Approach}
We propose an end-to-end network which combines both the global fingerprint image features and local patch based image features to obtain the final binary classification results. The whole procedure can be summarized in 6-steps. The first step involves preprocessing where the gray scale images of LiveDet datasets are converted into 3-channel images by adding LBP and Gabor feature channels. The second step is patch extraction followed by simultaneous training of two DenseNets for whole image and patch feature extractions. After the feature map extraction, channel, spatial and patch attention are performed on the patches. This is followed by the final step of patch feature fusion with the whole image and classification. An overview of the proposed model is shown in Figure 2.
\subsection{Preprocessing}
\begin{figure}[t]
\begin{center}
  \fbox{ \includegraphics[width = 0.7\linewidth]{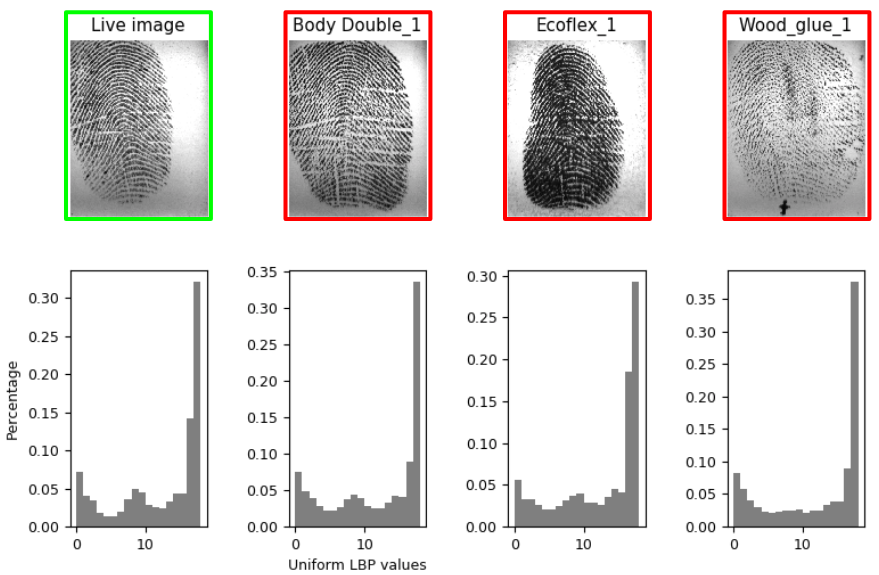}}
\end{center}
   \caption{LBP histograms for Live fingerprint and its corresponding fake counterparts for user 001\_1\_25 on Greenbit sensor from LiveDet 2017 dataset. }
\label{fig:LBP}
\end{figure}

As stated in the above section, we use preprocessing to convert the gray scale fingerprint sensor images to 3-channel images. As the fingerprints sensor images visually do not have any discriminative features, these additions facilitate the network to obtain more robust and better classification results on intra-sensor, inter-sensor, cross-material and cross-sensor datasets. We use two preprocessing methods, the first one is Local Binary Pattern and second is Gabor filters. 
\\\textbf{Local Binary Patterns(LBP)} are texture descriptors made popular by the work of Ojala \textit{et al.} \cite{Ojala1994PerformanceEO} in 1994. An LBP operator divides an image into cells of n$\times$n 
and a label is assigned to each cell after thresholding the neighboring pixel with the center pixel. The end result is an $2^n$-bit code representing all the $2^n$ possible combinations. As the comparison of the central pixel is made with the neighborhood pixels, LBP is an illumination invariant descriptor. We use an extension of this, called uniform or rotation-invariant LBP descriptor \cite{Ojala:2002:MGR:628329.628808}. This is used to reduce the length of feature vector as it contains only two bit transitions; 0 or 1, which in turn reduces the memory requirement. The rotation-invariant LBP histograms of live and it's corresponding fake fingerprints are shown in Figure \ref{fig:LBP}.

\begin{figure}[h]
\begin{center}
  \fbox{ \includegraphics[width = 0.7\linewidth]{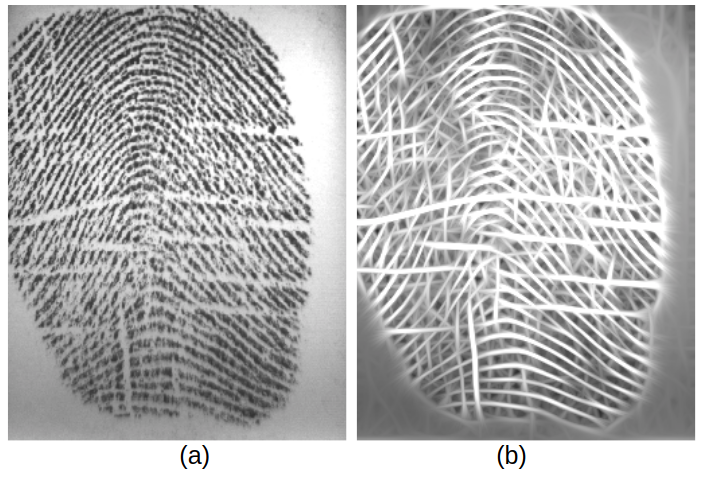}}
\end{center}
   \caption{(a) Original image , (b) Gabor filtered output for kernel size 51 and theta 11.55 degrees. }
\label{fig:Gabor}
\end{figure}

We create the third channel using the method of Gabor filtering.
\\\textbf{Gabor filter} is extensively used for various image processing applications like, edge detection, feature extraction, texture analysis etc. These filters can be viewed as special classes of band pass filters that have been known to possess optimal localization properties in both spatial and frequency domains. This makes them suitable for texture classification problems. We use a Gabor filter with kernel size of 51 and theta of 11.55 degrees to obtain the edges and texture intricacies of the fingerprint images. The output of Gabor filtered fingerprint is shown in the Figure \ref{fig:Gabor}.\\\textbf{Data Augmentation} is also added on the dataset to increase classifier robustness. We add standard augmentation methods like image resizing, random affine transforms and random crop on the data. These augmentation methods considerably reduce the memory requirements of the network without compromising its performance. 

\subsection{Implemented Network Details}
\textbf{Base Network:} Any of the standard network architectures like VGG \cite{Simonyan14verydeep}, ResNets \cite{He2015DeepRL}, Google Inception \cite{DBLP:journals/corr/SzegedyLJSRAEVR14}, MobileNet \cite{chollet2016xception}, DenseNets \cite{Huang2016Densely} etc. can be used to extract feature maps for both the whole image and the patches. We chose DenseNet as our backbone network because of its several compelling advantages like: (i) DenseNet has successfully obtained state-of=the-art results with high memory efficiency and less computation. (ii) It also has very few parameters for example; It has 0.8 million parameter for 100-layers with a growth rate k = 12 as compared to other networks like VGG which has 138 million parameter and ResNet-50 which has 25 million parameters. (iii) Due to the presence of densely connected layers, it alleviates the vanishing gradient problem and also strengthens feature propagation. (iv) It is easier to export this network on to hardware devices like mobile phones, FPGAs etc. due to it's small number of parameters. \\Our model consists of two standard DenseNets which are trained from scratch. We chose DenseNet with 40 layers and growth rate(k) 48 as network 1. This is obtained after several trails with different network parameters like DenseNet-121, DenseNet-169, DenseNet-190 etc.as it was optimal in terms to memory efficiency and accuracy. The second DenseNet consists of 10 layers and has a growth rate of  \textit{k = 6}. The results of Network-1 and Network-2 when trained on LivDet 2015 dataset separately are summarized in the Table 1.

\begin{table}[htb]

\resizebox{\columnwidth}{!}{%
\begin{tabular}{|c|c|c|c|}
\hline

\textbf{Training Dataset}         & \textbf{Testing Dataset}       & \textbf{Network-1 ACE}       & \textbf{Network-2 ACE}\\
\hline \hline
CrossMatch 2015  & CrossMatch 2015 & \textbf{0.21} &1.70\\
\hline
CrossMatch 2015 & Hi Scan 2015 & 3.67 & \textbf{2.13}\\
\hline
CrossMatch 2015 & GreenBit 2015 & 4.12 & \textbf{3.27}\\
\hline
CrossMatch 2015 & Digital Persona 2015 & 6.87 & \textbf{2.67}\\
\hline
GreenBit 2015 & GreenBit 2015 &\textbf{0.88} &2.75 \\
\hline
GreenBit 2015 & Hi Scan 2015 &\textbf{4.23} & \textbf{3.48} \\
\hline
GreenBit 2015 & Digital Persona 2015 &\textbf{5.75} & \textbf{2.87} \\
\hline
GreenBit 2015 & CrossMatch 2015 &\textbf{2.64} &3.72 \\
\hline
Digital Persona 2015 & Digital Persona 2015 &1.37 &\textbf{0.67} \\
\hline
Digital Persona 2015 & Hi Scan 2015 & 5.81 &\textbf{2.43} \\
\hline
Digital Persona 2015 & GreenBit 2015 &\textbf{3.52} & 4.72 \\
\hline
Digital Persona 2015 & CrossMatch 2015 &\textbf{2.66} & 5.43 \\
\hline

\end{tabular}
}
\caption{Comparison of intra sensor and cross sensor ACE between Network-1 and Network-2 for LivDet 2015 dataset}
\label{1234}
\end{table}

\begin{table*}[!htpb]
\begin{flushleft}

\vspace{-1.32cm}
\scalebox{0.86}{
\resizebox{1.20\textwidth}{0.62\textheight}{%
\large
\begin{tabular}{|c|c|c|c|c|c|c|}
\hline

\multirow{2}{*}{\textbf{Dataset Year}}& \multirow{2}{*}{\textbf{Training Sensor}}& \multirow{2}{*}{\textbf{Spoof Materials used for training}} & \multirow{2}{*}{\textbf{Testing Sensor}} & \multirow{2}{*}{\textbf{Spoof Materials used for testing}} &\multicolumn{2}{c|}{\textbf{ACE =    ((FPR + FNR) / 2)*100 \%}}\\

\cline{6-7}
    &&&&& \textbf{Current S.O.T.A} & \textbf{Proposed Network}\\
\hline \hline
\multirow{9}{*}{\textbf{LiveDet 2017}} & \multirow{3}{*}{DigitalPersona U.are.U 5160} & \multirow{3}{*}{ Wood Glue, Ecoflex and Body Double} & DigitalPersona U.are.U 5160 & Latex,Gelatine, Liquid Ecoflex  & $-$ & \textbf{\color{blue}0.71}\\
\cline{4-7}
&&& GreenBit DactyScan84c & Latex,Gelatine, Liquid Ecoflex  &  $-$ & \textbf{\color{blue}2.19}\\
\cline{4-7}
&&& Orcanthus Certis2 Imag & Latex,Gelatine, Liquid Ecoflex & $-$ & \textbf{\color{blue}3.21}\\
\cline{2-7}
& \multirow{3}{*}{GreenBit DactyScan84c} & \multirow{3}{*}{ Wood Glue, Ecoflex and Body Double} & GreenBit DactyScan84c & Latex,Gelatine, Liquid Ecoflex  & $-$ & \textbf{\color{blue}0.68}\\
\cline{4-7}

&&& Orcanthus Certis2 Imag & Latex,Gelatine, Liquid Ecoflex   & $-$ & \textbf{\color{blue}2.73}\\
\cline{4-7}
&&& DigitalPersona U.are.U 5160 & Latex,Gelatine, Liquid Ecoflex  & $-$ & \textbf{\color{blue}5.32}\\
\cline{2-7}
& \multirow{3}{*}{Orcanthus Certis2 Imag} & \multirow{3}{*}{Wood Glue, Ecoflex and Body Double} & Orcanthus Certis2 Imag & Latex,Gelatine, Liquid Ecoflex  & $-$ & \textbf{\color{blue}0.03}\\
\cline{4-7}

&&& GreenBit DactyScan84c &  Latex,Gelatine, Liquid Ecoflex & $-$ & \textbf{\color{blue}6.77}\\
\cline{4-7}
&&& DigitalPersona U.are.U 5160 & Latex,Gelatine, Liquid Ecoflex & $-$ & \textbf{\color{blue}7.46}\\
\hline \hline
\multicolumn{5}{| c |}{\textbf{Average}}& NA &\color{magenta}\textbf{0.48}\\
\hline \hline
\multirow{21}{*}{\textbf{LiveDet 2015}} & \multirow{6}{*}{Cross-Match  L  SCAN  GUARDIAN} & \multirow{6}{*}{ Body Double,Ecoflex,Playdoh} & Cross-Match  L  SCAN  GUARDIAN & Body Double,Ecoflex,Playdoh  &1.525\cite{Chugh2017FingerprintSB} & \textbf{\color{blue}0.23}\\
\cline{4-7}
&&&  Cross-Match  L  SCAN  GUARDIAN & Gelatin, OOMOO  &2.475\cite{Chugh2017FingerprintSB} & \textbf{\color{blue}0.37}\\
\cline{4-7}
&&&  DigitalPersona U.are.U 5160 & Ecoflex, Gelatin, Latex, Wood Glue  &$-$ &\textbf{\color{blue}1.12} \\
\cline{4-7}
&&&  GreenBit DactyScan84c & Ecoflex, Gelatin, Latex, Wood Glue &$-$ & \textbf{\color{blue}1.97}\\
\cline{4-7}
&&&  HiScan-PRO & Ecoflex, Gelatin, Latex, Wood Glue &$-$ & \textbf{\color{blue}1.78}\\
\cline{4-7}
&&&  Cross-Match  L  SCAN  GUARDIAN - 2013 & Ecoflex, Gelatin, Latex, Wood Glue &$-$ & \textbf{\color{blue}1.25}\\
\cline{2-7}

& \multirow{5}{*}{HiScan-PRO} & \multirow{5}{*}{Ecoflex, Gelatin, Latex, Wood Glue}& HiScan-PRO & Ecoflex, Gelatin, Latex, Wood Glue  &$-$ &\textbf{\color{blue}0.63}\\
\cline{4-7}
&&&  HiScan-PRO & Liquid$-$Ecoflex,RTV  &$-$ & \textbf{\color{blue}2.57}\\
\cline{4-7}
&&&  DigitalPersona U.are.U 5160 & Ecoflex, Gelatin, Latex, Wood Glue  &$-$ & \textbf{\color{blue}1.87}\\
\cline{4-7}
&&&  GreenBit DactyScan84c & Ecoflex, Gelatin, Latex, Wood Glue &$-$ &\textbf{\color{blue}2.05}\\
\cline{4-7}
&&&  Cross-Match  L  SCAN  GUARDIAN & Body Double,Ecoflex,Playdoh &$-$ & \textbf{\color{blue}2.31}\\
\cline{2-7}
& \multirow{5}{*}{GreenBit DactyScan84c} & \multirow{5}{*}{Ecoflex, Gelatin, Latex, Wood Glue}&GreenBit DactyScan84c & Ecoflex, Gelatin, Latex, Wood Glue &3.9\cite{Chugh2017FingerprintSB} & \textbf{\color{blue}1.81}\\
\cline{4-7}
&&& GreenBit DactyScan84c & Liquid$-$Ecoflex,RTV &5.65\cite{Chugh2017FingerprintSB} & \textbf{\color{blue}2.82}\\
\cline{4-7}
&&&  DigitalPersona U.are.U 5160 & Ecoflex, Gelatin, Latex, Wood Glue & $-$  & \textbf{\color{blue} 2.51}\\
\cline{4-7}
&&&  HiScan-PRO & Ecoflex, Gelatin, Latex, Wood Glue & $-$ &\textbf{\color{blue}2.09}\\
\cline{4-7}
&&&  Cross-Match  L  SCAN  GUARDIAN & Body Double,Ecoflex,Playdoh & $-$ & \textbf{\color{blue}2.35}\\
\cline{2-7}
& \multirow{5}{*}{DigitalPersona U.are.U 5160} & \multirow{5}{*}{Ecoflex, Gelatin, Latex, Wood Glue}& DigitalPersona U.are.U 5160 & Ecoflex, Gelatin, Latex, Wood Glue & 7.85\cite{Chugh2017FingerprintSB}  & \textbf{\color{blue}1.72}\\
\cline{4-7}
&&&  DigitalPersona U.are.U 5160 & Liquid$-$Ecoflex,RTV & 7.05\cite{Chugh2017FingerprintSB}  & \textbf{\color{blue}2.63}\\
\cline{4-7}
&&&  GreenBit DactyScan84c & Ecoflex, Gelatin, Latex, Wood Glue & $-$  & \textbf{\color{blue}4.67}\\
\cline{4-7}
&&&  HiScan-PRO & Ecoflex, Gelatin, Latex, Wood Glue & $-$  & \textbf{\color{blue}2.94}\\
\cline{4-7}
&&&  Cross-Match  L  SCAN  GUARDIAN & Body Double,Ecoflex,Playdoh & $-$  & \textbf{\color{blue}3.95}\\
\hline \hline
\multicolumn{5}{| c |}{\textbf{Average}}&0.97$^*$\cite{Chugh2017FingerprintSB}&\color{magenta}\textbf{0.84}$^*$\\
\hline \hline
\multirow{14}{*}{\textbf{LiveDet 2013}} & \multirow{4}{*}{Biometrika  FX2000} &  \multirow{4}{*}{Ecoflex, ,Gelatin, Modasil, Latex, Wood Glue} & Biometrika  FX2000 & Ecoflex, ,Gelatin, Modasil, Latex, Wood Glue & \textbf{\color{blue}0.20\cite{Chugh2017FingerprintSB}}  & 0.24\\
\cline{4-7}
&&& Biometrika  FX2000 -2011 & Ecoflex, ,Gelatin, Modasil, Latex, Wood Glue & 31.16\cite{Chugh2017FingerprintSB}  & \textbf{\color{blue}11.90}\\
\cline{4-7}
&&& Italdata ET10 & Ecoflex, ,Gelatin, Modasil, Latex, Wood Glue & \textbf{1.5\cite{Pala2017}}  & 6.46\\
\cline{4-7}
&&& Crossmatch L SCAN GUARDIAN & Body Double,Ecoflex,Playdoh & $-$  & \textbf{\color{blue}2.32}\\
\cline{2-7}
&\multirow{4}{*}{Italdata ET10} & \multirow{4}{*}{Ecoflex, ,Gelatin, Modasil, Latex, Wood Glue} & Italdata ET10 & Ecoflex, ,Gelatin, Modasil, Latex, Wood Glue & \textbf{\color{blue}0.30\cite{Chugh2017FingerprintSB}} & 0.32\\
\cline{4-7}
&&& Crossmatch L SCAN GUARDIAN & Body Double,Ecoflex,Playdoh & $-$  & \textbf{\color{blue}1.35}\\
\cline{4-7}
&&& Biometrika  FX2000 & Ecoflex, ,Gelatin, Modasil, Latex, Wood Glue &2.30\cite{article} & \textbf{\color{blue}1.75}\\
\cline{2-7}
&\multirow{4}{*}{Crossmatch L SCAN GUARDIAN} & \multirow{4}{*}{Body Double,Ecoflex,Playdoh} &Crossmatch L SCAN GUARDIAN & Body Double,Ecoflex,Playdoh & $-$ & \textbf{\color{blue}0.34}\\
\cline{4-7}
&&& Crossmatch L SCAN GUARDIAN-2015 & Body Double,Ecoflex,Playdoh & $-$ & \textbf{\color{blue}1.58}\\
\cline{4-7}
&&& Italdata ET10 & Ecoflex, ,Gelatin, Modasil, Latex, Wood Glue & $-$ & \textbf{\color{blue}2.47}\\
\cline{4-7}
&&& Biometrika  FX2000 & Ecoflex, ,Gelatin, Modasil, Latex, Wood Glue & $-$ & \textbf{\color{blue}2.65}\\
\hline \hline
\multicolumn{5}{| c |}{\textbf{Average}}&\color{magenta}\textbf{0.25}$^*$&\textbf{0.28}$^*$\\
\hline \hline
\multirow{17}{*}{\textbf{LiveDet 2011}} & \multirow{4}{*}{Digital 4000B} & \multirow{4}{*}{Gelatin, Latex, Playdoh, Silicone, WoodGlue} &Digital 4000B & Gelatin, Latex, Playdoh, Silicone, WoodGlue & \textbf{\color{blue}1.61\cite{Chugh2017FingerprintSB}} & 2.43\\
\cline{4-7}
&&& Biometrika  FX2000 & Ecoflex, Silgum, Gelatin, WoodGlue, Latex & $-$ &\textbf{\color{blue}6.21}\\
\cline{4-7}
&&& Italdata ET10 & Ecoflex, Gelatin, Latex, Silgum, WoodGlue & $-$ & \textbf{\color{blue}5.17}\\
\cline{4-7}
&&& Sagem MSO300 & Gelatin, Latex, Playdoh, Silicone, Wood Glue & $-$ & \textbf{\color{blue}11.89}\\
\cline{2-7}
& \multirow{4}{*}{Biometrika  FX2000} & \multirow{4}{*}{Ecoflex, Silgum, Gelatin, WoodGlue, Latex} &Biometrika  FX2000 & Ecoflex, ,Gelatin, Modasil, Latex, Wood Glue & 1.24\cite{Chugh2017FingerprintSB} & \textbf{\color{blue}0.19}\\
\cline{4-7}
&&& Biometrika  FX2000 & Ecoflex, Silgum, Gelatin, WoodGlue, Latex & 7.60\cite{Chugh2017FingerprintSB} &\textbf{\color{blue}2.18}\\
\cline{4-7}
&&& Italdata ET10 & Ecoflex, Gelatin, Latex, Silgum, WoodGlue & 25.35\cite{Chugh2017FingerprintSB} & \textbf{\color{blue}2.13}\\
\cline{4-7}
&&& Sagem MSO300 & Gelatin, Latex, Playdoh, Silicone, Wood Glue & $-$ & \textbf{\color{blue}2.43}\\
\cline{4-7}
&&& Digital 4000B & Gelatin, Latex, Playdoh, Silicone, WoodGlue & $-$ & \textbf{\color{blue}3.48}\\
\cline{2-7}
& \multirow{4}{*}{Sagem MSO300} & \multirow{4}{*}{Gelatin, Latex, Playdoh, Silicone, Wood Glue} & Sagem MSO300 & Gelatin, Latex, Playdoh, Silicone, Wood Glue & 1.23\cite{Pala2017} &\textbf{\color{blue}0.96}\\
\cline{4-7}
&&& Italdata ET10 & Ecoflex, Gelatin, Latex, Silgum, WoodGlue & $-$ & \textbf{\color{blue}2.46}\\
\cline{4-7}
&&& Biometrika  FX2000 & Ecoflex, Silgum, Gelatin, WoodGlue, Latex & $-$ &\textbf{\color{blue}6.78}\\
\cline{4-7}
&&&Digital 4000B & Gelatin, Latex, Playdoh, Silicone, WoodGlue & $-$ & \textbf{\color{blue}11.35}\\
\cline{2-7}
& \multirow{4}{*}{Italdata ET10} & \multirow{4}{*}{Ecoflex, Gelatin, Latex, Silgum, WoodGlue} & Italdata ET10 & Ecoflex, Gelatin, Latex, Silgum, WoodGlue & 2.45\cite{Chugh2017FingerprintSB}  & \textbf{\color{blue}1.06}\\
\cline{4-7}
&&& Italdata ET10 - 2013 & Ecoflex, ,Gelatin, Modasil, Latex, Wood Glue & 6.70\cite{Chugh2017FingerprintSB} & \textbf{\color{blue}2.54}\\
\cline{4-7}
&&& Biometrika  FX2000 & Ecoflex, Silgum, Gelatin, WoodGlue, Latex & 25.21\cite{Chugh2017FingerprintSB} &\textbf{\color{blue}9.90}\\
\cline{4-7}
&&&Digital 4000B & Gelatin, Latex, Playdoh, Silicone, WoodGlue & $-$ & \textbf{\color{blue}13.47}\\
\cline{4-7}
&&& Sagem MSO300 & Gelatin, Latex, Playdoh, Silicone, Wood Glue & $-$ & \textbf{\color{blue}5.93}\\
\hline \hline
\multicolumn{5}{|c|}{\textbf{Average}}&1.63$^*$ & \textbf{\color{magenta}1.16$^*$}
\\
\hline \hline
\end{tabular}}}
\end{flushleft}
\label{6}
\caption{The overall performance comparison between different sensors across different datasets. $^*$ The average ACE takes only the intra sensor errors into consideration and for LivDet 2013, only Italdata and Biometrika are considered for comparison purpose only.}
\end{table*}

\textbf{Attention Module:} We propose a novel patch attention model along with channel and spatial attention networks which identifies the most discriminative patches amongst the \textit{n} given patches and allocates corresponding weights to each patch.
As we can see from the overview of the complete proposed model in Figure 2, once feature maps are obtained from each patch after training it with the second DenseNet (10 layer, growth rate = 6), they are passed through a channel attention network.
\\\textbf{Channel Attention Network} \cite{Woo_2018_ECCV} is used to obtain the channel attention map by exploiting the inter channel relationship. Using channel attention network we get \textbf{what} is the most important feature of an image. For obtaining this we compress the feature map in spatial dimensions ($H \times W$) using average pooling and max pooling techniques. Let the feature vector at the end of each patch be defined as \textit{f}, and let the image dimensions be $(C \times H \times W)$. The feature vector obtained after average pooling and max pooling can be given as $f_{avg}$ and $f_{max}$ respectively. These feature maps are forwarded to a shared layer in the channel attention network which is a multi-layer perceptron with one hidden layer. The hidden layer reduces the channel parameters by a defined reduction ratio $r$. The operation of the channel attention model is mathematically summarized in equation (1).
\begin{equation}\label{1}
	F^c = \sigma(MLP(f_{avg})+ MLP(f_{max}))
\end{equation}
where $\sigma $ denotes the sigmoid function and $F^c$ denotes channel attention output.
This channel attention is followed by a spatial attention module with the output of channel attention network is given as an input to it.
\\\textbf{Spatial Attention Network} \cite{Woo_2018_ECCV} highlights \textbf{where} the important information exists in the the patch. This network extracts the most informative regions in the feature map using inter-spatial relationship. Like channel attention, to obtain spatial attention, the channel information is restored and compressed by average pooling ($F^c_{avg}$) and max pooling ($F^c_{max}$) across the channel dimension. These generate two 2-D maps of dimensions $(1 \times H \times W)$. The two 2-D feature maps are concatenated and convoluted with a standard convolution layer to obtain a 2-D spatial attention map. The operation of the spatial attention model is mathematically summarized in equation (2).
\begin{equation}\label{2}
	F^s = \sigma(L([F^c_{avg};F^c_{max}]))
\end{equation}
where, $L$ is a filter, $\sigma$ denotes the sigmoid function and $F^s$ denotes spatial attention output.
The output of spatial attention model is the input for the patch attention network. 
\\\textbf{Patch Attention Network} is a novel attention module proposed in this paper, which highlights the most discriminative patches among the $n$ patches in the network. The input for the patch attention model is a concatenated output map of all the maps obtained after channel attention and spatial attention modules. Therefore, the dimension of the final input to the patch attention model is ${n \times C \times H \times W}$ where, n is the number of patches. Similar to the earlier attention network, the features of both the channel attention and the spatial attention are aggregated using two pooling operations. This ensures that the vital information obtained from the channel and spatial attention modules is preserved. We first apply average pooling across spatial dimension which gives us a map $M_s$ of dimension ${n \times C}$. After average pooling across spatial map, average pooling across channel map is performed which gives an output map $M_c$ of dimension  ${n \times 1}$. It is then passed through a multi-layer perceptron with one hidden layer. The hidden layer reduces the patch parameters by a given reduction ratio $r$. The operation of the patch attention network is mathematically shown in equation(3) and equation(4).

\begin{equation}\label{3}
	F^p = \sigma(MLP(M_c(M_s(F^s))))
\end{equation}
In terms of weights this can be given as: 
\begin{equation}\label{4}
	F_i^p = \sigma(W_i(M_c))
\end{equation}
Where, $\sigma$ denotes sigmoid, and $W_i$ is the weight of $i^{th}$ patch such that  $W_i$ $\in$ $R$
\section{Experimental Results}

\subsection{Implementation Details}
We train our model from scratch and implement it with a batch size of $n + 1$. Each fingerprint is of size $224 \times 224$ from which $n$ patches of size  $56 \times 56$ are extracted and trained on Network-2 which is a 10-layer DenseNet with a growth rate of $k = 6$. The whole fingerprint image is trained on Network-1 which is a 40 layer DenseNet with a growth rate of $k = 48$. Standard cross-entropy loss is used for training both the networks. To optimize the loss, we use the SGD optimizer with nesterov momentum \cite{Qian:1999:MTG:307343.307376} with a learning rate initialized to 0.006 and a weight decay of 1e-4. The network is trained for 500 epochs with a momentum initialized at 0.9.
Our model is implemented on Pytorch platform and has 2.74M parameters. It is implemented on GeForce RTX 2080 Ti GPU. All our experiments on different datasets follow the same setting as above.
\subsection{Evaluation Metric}
The proposed approach is evaluated using the performance evaluation metrics used for LiveDet\cite{2017:RFL:3063814.3063826}
The following metrics are evaluated:
\\$F_{errlive}$: Percentage of misclassified live fingerprints
\\$F_{errfake}$: Percentage of misclassified fake fingerprints
\\The Average Classification Error (ACE) is given as:
\begin{equation}
    ACE = \frac{F_{errlive} + F_{errfake}}{2}
\end{equation}
\subsection{Datasets}
The proposed approach is trained and tested on four datasets provided by the Liveness Detection Competition (LivDet) in the years of 2011 \cite{LiveDet_2011}, 2013 \cite{LivDet_2013} 2015 \cite{LiveDet_2015} and 2017 \cite{LivDet_2017}

\textbf{LivDet 2011} comprises 16,000 images obtained using four
different sensors: Biometrika FX2000, Digital 4000B, Italdata
ET10, and Sagem MSO300, each having 2000 images of fake
and real fingerprints. 

\textbf{LivDet 2013} comprises 16,000 images acquired from four
different sensors: Biometrika FX2000, Crossmatch L SCAN
GUARDIAN, Italdata ET10, and Swipe, each having approximately 2,000 images of fake and real fingerprints. 

\textbf{LivDet 2015} comprises of 19,431 fingerprint images acquired from four different sensors: CrossMatch L SCAN GUARDIAN, Digital Persona U are U 5160, HiScan-PRO, GreenBit DactyScan26 each having approximately 1000 fingerprints for training and 1000 for testing. 8983 fingerprints both Live and Fake are used for training and 10,448 are used for testing purposes. 

\textbf{LivDet 2017} comprises of 17,820 fingerprint images acquired from three different sensors:Digital Persona U are U 5160, Orcathus Certis2 Image, GreenBit DactyScan2, each having approximately 2200 fingerprints for training and 3740 for testing. For each of the above mentioned fingerprint dataset, we evaluate the same-material ACE, cross-material ACE, cross-sensor ACE and cross-dataset ACE. The complete evaluation is summarized in Table 2.
\subsection{Analysis}
An exhaustive experimental analysis is provided for the four datasets LivDet 2011, 2013, 2015 and 2017 which is summarized in Table 2. For each sensor in the dataset, same-material, cross-material, cross-sensor and cross-dataset error metrics are obtained.

\textbf{Intra-Sensor error} metric is obtained when the network is trained and tested on the same dataset with same spoof materials(same-material) or different spoof materials(cross-material). We can see from Table 2. that for almost all datasets there is a considerable reduction in error when tested on both same material and different material. Our proposed method obtained better intra-sensor accuracies for almost all datasets. There is a striking \textbf{85\%} decrease in the average classification error when trained and tested on same spoof materials and also different spoof materials for CrossMatch 2015 dataset when compared with the previous state-of-the-art approaches. This can be attributed to the presence of both global features and local features learned by our network. If we compare the same material average classification error with the results given by Network-1 and Network-2 in Table 1 we can see that with only preprocessing and DenseNet, Network-1 outperforms the state-of-the-art.
Cross Material error  is considerably reduced due to the addition of Network-2. The 10-layer DenseNet in Network-2 which is trained on patches, learns the local features which along with attention network improves the cross-material accuracy.
The Digital Persona fingerprints are of smaller resolution as compared the other sensors but, this did not pose a problem with our method as we include both local patch features and the global features. The average classification error is significantly lower with our method as compared to the previous state-of-the-art methods. The plots for intra-sensor ACE is shown in Figure 5.

\begin{figure}[h!]
 \begin{flushleft}
 \hspace{-0.8cm}
  \includegraphics[scale = 0.40]{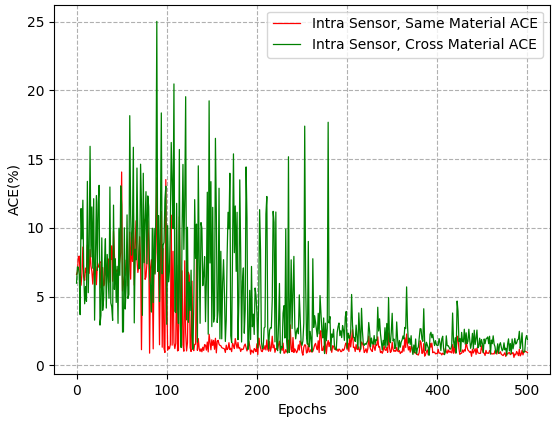}
  \end{flushleft}
  \caption{Intra-Sensor average classification error for network  trained on CrossMatch L SCAN GUARDIAN 2015 sensor and tested on data having same testing spoof materials and different testing spoof materials.} 
\end{figure}

\textbf{Cross-Sensor error} metric is obtained when the training images belong to one sensor and the testing images belong to a different sensor of the same dataset. We can infer from the summary table that our approach provides exemplary results in case of cross-sensor metric also. If we compare the cross-sensor results of LivDet 2015 dataset. The cross-sensor ACE for CrossMatch sensor when tested with GreenBit, Digital Persona and Hi Scan data is 1.97, 1.12, 1.78 respectively which is relatively lower compared to that of previous state-of-the-art approaches. This is due the innate property of our network to learn common characteristics among the datasets which help in the subsequent classification. The cross-sensor metrics outperforms the state of the art for all datasets except for LivDet 2013. Figure 6. shows the average classification error plot for our network trained on CrossMatch sensor and tested on GreenBit, Hi Scan and Digital Persona.

\begin{figure}[t]\label{fig.6.}   
\centering
 \begin{flushleft}
 \hspace{-0.8cm}
  \includegraphics[scale = 0.40]{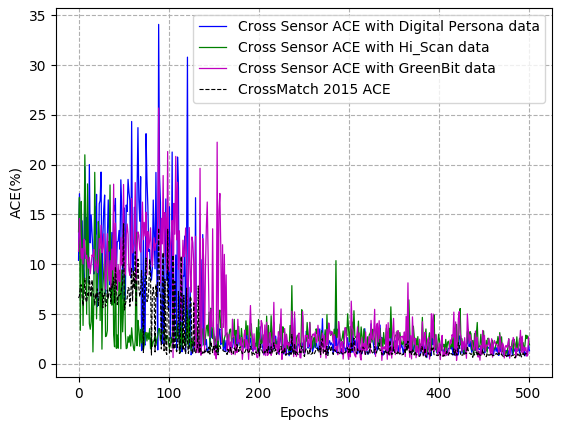}
  \end{flushleft}
  \caption{Cross-Sensor average classification error for network trained on CrossMatch L SCAN GUARDIAN 2015 sensor and tested with the data of Hi Scan, GreenBit and Digital Persona sensors.}
\end{figure}

\textbf{Cross-Dataset error} metric is obtained when the training images belong to the same sensor of one dataset and testing images belong to same sensor of different datasets. For example, network is trained on Biometrika sensor from LivDet 2011 and tested on Biometrika sensor data from LivDet 2013. 
As our network adeptly learns the common characteristics to classify a fingerprint as live or fake, we obtain considerably good cross dataset results. If we compare the cross dataset result of Italdata 2011 when tested with Italdata 2013, we can see that there is a \textbf{60\%} reduction in the average classification error. Even though the cross sensor and cross dataset classification errors are significantly lower with our approach, there is still a considerable amount of scope of improvement in this area. Figure 7. shows the cross dataset ACE for CrossMatch 2015.

\begin{figure}[ht!]\label{fig.7.}   
 \begin{flushleft}
 \hspace{-0.51cm}
  \includegraphics[scale = 0.40]{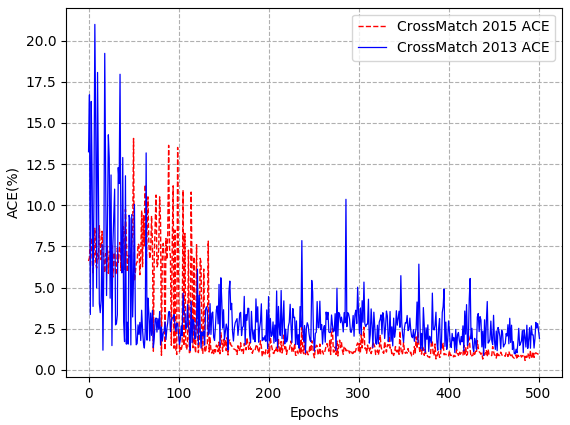}
  \end{flushleft}
  \caption{Cross-Dataset average classification error for network trained on CrossMatch 2015 dataset and tested on CrossMatch 2013 dataset.}
\end{figure}
\section{Conclusion}
In this paper, we propose a novel end-to-end fingerprint spoof detection network. The proposed network automatically extracts informative patches using a novel patch based attention mechanism. Use of DenseNet as the base network optimizes memory requirement. Furthermore, use of global fingerprint image along with the fingerprint patches helps in improving network robustness and generalizes it's performance across cross sensor, cross material and cross dataset. The effectiveness of the proposed network is validated through extensive set of experiments carried over various datasets, where a significant improvement from the current state-of-the-art is achieved.
In future, we plan to integrate this proposed network with fingerprint authentication and recognition networks to obtain a complete fingerprint recognition system with inbuilt spoof detector.
\section{Acknowledgement}
This research work is supported and funded by the Department of Science and Technology(DST) under Indo-Korean bilateral scheme.
{\small
\bibliographystyle{ieee}
\bibliography{egbib}
}
\end{document}